\documentclass[conference]{IEEEtran}
\IEEEoverridecommandlockouts
\usepackage{cite}
\usepackage{amsmath,amssymb,amsfonts}
\usepackage{algorithmic}
\usepackage{graphicx}
\usepackage{textcomp}
\usepackage{xcolor}
\def\BibTeX{{\rm B\kern-.05em{\sc i\kern-.025em b}\kern-.08em
    T\kern-.1667em\lower.7ex\hbox{E}\kern-.125emX}}

\begin{document}

\title{A Survey on Mathematical Aspects \\ of Machine Learning in GeoPhysics: \\ The Cases of Weather Forecast, Wind Energy,\\ Wave Energy, Oil and Gas Exploration \\
}

\author{\IEEEauthorblockN{1\textsuperscript{st} Miroslav Kosanić}
\IEEEauthorblockA{
\textit{School of Electrical Engineering}\\
Belgrade, Serbia \\
kosanic.miroslav@gmail.com}
\and
\IEEEauthorblockN{2\textsuperscript{nd} Veljko Milutinović}
\IEEEauthorblockA{\textit{Indiana University Bloomington} \\
\textit{IEEE Life Fellow}\\
Indiana, USA \\
vm@etf.rs}
}

\maketitle

\begin{abstract}
This paper reviews the most notable works applying machine learning techniques (ML) in the context of geophysics and corresponding subbranches. We showcase both the progress achieved to date as well as the important future directions for further research, while providing an adequate background in the fields of weather forecast, wind energy, wave energy, oil and gas exploration. The objective is to reflect on the previous successes and provide a comprehensive review of the synergy between these two fields in order to speed up the novel approaches of machine learning techniques in geophysics. Last but not least, we would like to point out possible improvements, some of which are related to implementation of ML algorithms using DataFlow paradigm as a means of performance acceleration.

\end{abstract}


\section{Introduction}
The second half of the 20th century has seen a lot of scientific advances, particularly in the field of computer science. Foundations for Machine learning were set during 40s and 50s where the notion of artificial intelligence (\textbf{AI}) was firstly introduced by Turing, and subsequently ML was defined by Arthur Samuel (1959) and Tom Mitchell (1997). ML nowadays has become necessity in almost every science field. These recent advances were made possible due to increase, availability and advances in computer power, interconnected sensors and devices, and big data centers.

Geophysics emerged as a separate discipline during the 19th century, from the intersection of physical geography, geology, astronomy, meteorology, and physics.\cite{b0}. Nowadays, it is one of the most important fields with respect to new energy resources identification, classification and management. Machine learning and supercomputers are the tools of contemporary geophysics which spans from climate change projections to earthquake simulations and energy resources optimization.

The first applications of machine learning were seen during 1990s (\cite{b1}, \cite{b2}) after long AI winter and revival of the ML. Recent advances of Deep learning (DL) have allowed for a more suitable algorithms (related to spatio-temporal features of the data) in a data rich field such as the geophysics (\cite{b3}, \cite{b4}).

This paper aims to show successful synergy between the geophysics and ML communities and thus provide directions for further work both by industry and academia, so that the practical adoption of machine learning techniques for geophysical applications is further accelerated. In order to do so, we will analyse the recent machine learning applications for weather forecast, wind energy, wave energy, oil and gas exploration for previous 5 years. The main motivation for this work comes from the fact that most of the ML algorithms applied to big data are best implemented in the DataFlow paradigm. Our aim is to make unique contribution as the previous efforts (\cite{b59}, \cite{b60}, \cite{b61}, \cite{b62}, \cite{b63}, \cite{b64}) have focused separately on fields mentioned in the title of the paper and ML applications respectively, while we strive for more holistic and integrative approach of machine learning applied to forecasting of these energy resources. We focus on showcasing both the progress achieved to date as well as the important future directions for further research as we will try to identify performance bottlenecks and suggest suitable and promising solutions. As we would like to address audiences from both communities, we will briefly provide an adequate background in the fields of machine learning and of geophysics.

The rest of this paper is organised as follows: sections II and III will present the required background in geophysics and in machine learning; section IV provides statistics about the publications of ML applied to geophysics since the year 2015, while section V reviews published works over the last 5 years (2015-2020). Section VI discusses research directions for geophysics ML applications, and integration of data-driven and physics-based approaches due to the results that clearly show that their combination outperforms either approach in isolation.

\section{Machine learning basics}

\begin{figure}[htbp]
\centerline{\includegraphics[scale=0.2]{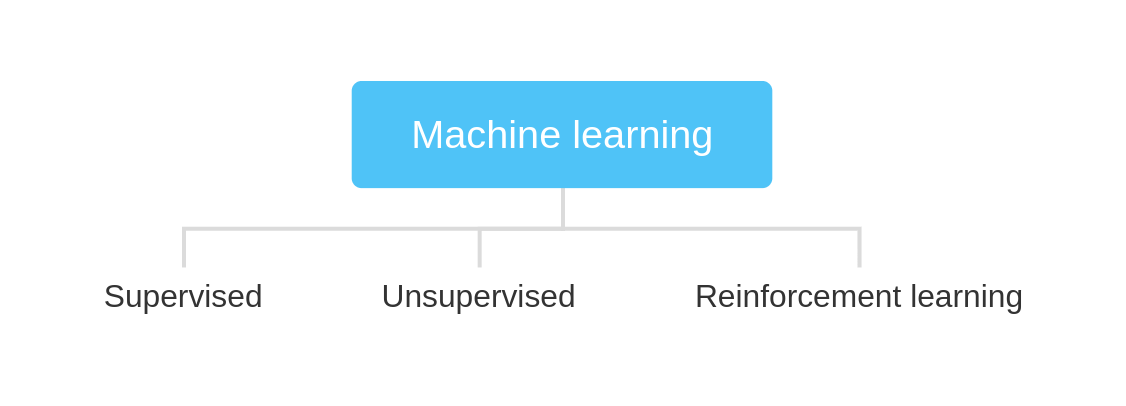}}
\caption{ML algorithms classification}
\label{fig1}
\end{figure}

As we can see in figure \ref{fig1}, a diverse array of ML algorithms can be divided into three categories, namely:
\begin{itemize}
    \item supervised
    \item unsupervised
    \item reinforcement
\end{itemize}

Supervised learning systems generally form
their predictions or conclusions via a learned mapping $f(x)$ which is based on training data $(x,y)$ and produces an output $y$ for each input $x$ (or a probability distribution over $y$ given $x$) \cite{b20}. Various forms of such mappings $f$ exist, like linear and nonlinear regression (\cite{ml1}, \cite{ml2}, \cite{ml3}, \cite{ml4}, \cite{ml5}), logistic regression ( \cite{ml6}, \cite{ml7}, \cite{ml8}),  decision trees (\cite{b21}, \cite{b22}, \cite{b22}, \cite{b23}, \cite{b24}, \cite{b25}, \cite{b26}) , decision forests ( \cite{b27}, \cite{b28}, \cite{b29}, \cite{b30}, \cite{b31}), support vector machines (\cite{b35}, \cite{b36}, \cite{b37}, \cite{b38}), neural networks ( \cite{b39}, \cite{b40}, \cite{b41}, \cite{b42}, \cite{b43}, \cite{b44}), kernel machines (\cite{b45}, \cite{b46}), and Bayesian classifiers and regressors (\cite{b48}, \cite{b49}, \cite{b50}).

The second group, unsupervised learning represents algorithms which analyse unlabeled data under assumptions about structural properties of the
data (e.g., algebraic, combinatorial, or probabilistic). If we assume that data lies on a low-dimensional manifold then algorithm's aim is to identify that manifold explicitly from data (\cite{b51}, \cite{b52}, \cite{b53}, \cite{ml31}).

Third group, reinforcement learning, spans algorithms which instead of having training examples that indicate the correct output for a given input, operate with the training data that provides only an indication as to whether an action is correct or not or how much of reward the algorithm gets by taking that particular action. If an action is incorrect or so to say "bad", the algorithms recalculates the probability distribution with regard to a set of possible actions and proceeds to find the correct action or action with the highest yield in terms of a reward in that state (\cite{b54}, \cite{b55}, \cite{b56}, \cite{b57}, \cite{b58}).

We will mainly focus on the first two groups and provide relevant examples in section V.

\section{Geophysics basics}

Geophysics is the field which deals with physics of the Earth and its atmosphere. The principal subdivisions of geophysics are \cite{b11}:
\begin{itemize}
    \item seismology
    \item geothermometry
    \item hydrology
    \item physical oceanography
    \item meteorology
    \item terrestrial magnetism
    \item gravity and geodesy
    \item atmospheric electricity and terrestrial magnetism
    \item tectonophysics
    \item exploration, engineering, and environmental geophysics
\end{itemize}

\begin{figure}[htbp]
\centerline{\includegraphics[scale=0.13]{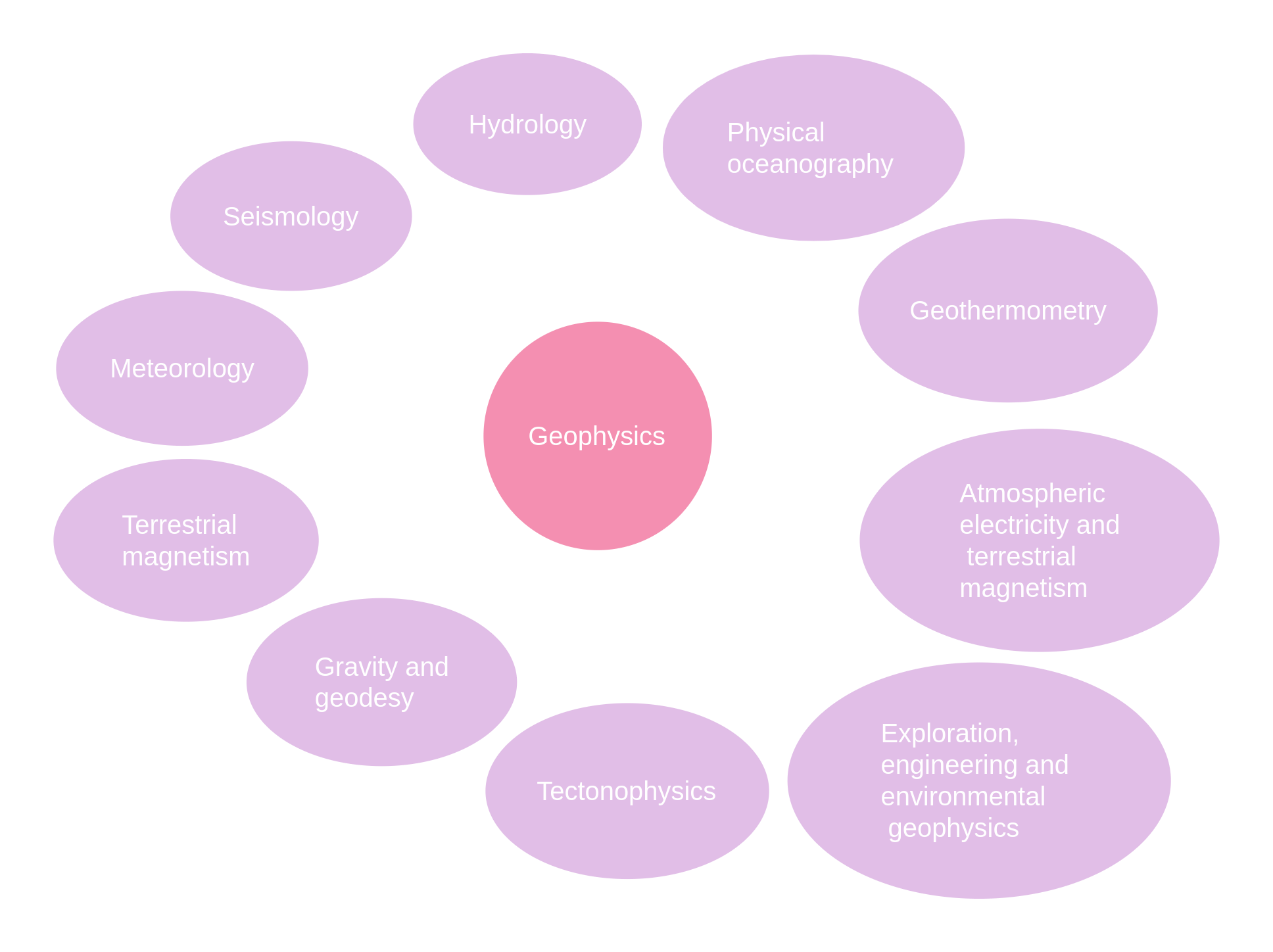}}
\caption{Geophysics subdivisions}
\label{fig2}
\end{figure}

Out of these, we will mainly focus on hydrology, physical oceanography, meteorology and exploration geophysics, thus we'll give their definitions. 

Hydrology as its names suggests is the study of the global water cycle and the physical, chemical, and biological processes involved in the different reservoirs and fluxes of water within this cycle \cite{b12}.

Physical oceanography addresses 
the ocean through both observations and
complex numerical model output used to quantitatively describe the fluid motions \cite{b13}.

Meteorology is the study of the atmosphere and its phenomena meaning every weather event or condition at any particular time and place that occur at the earths surface \cite{b14}. 

Exploration geophysics main application is related to prospecting for natural resources, mostly related to oil industry in search for hydrocarbons \cite{b15}.

The forecasting of weather (by which we will mainly refere to solar energy), wind and wave energy is important as to provide guidance for the electric power operation and power transmission system and to enhance the efficiency of energy capture and conversion. But as much as we would like to harvest only renewable sources of energy, oil and gas still remain big accelerators of industry progress, so their exploration is still key factor in energy sector.

In the literature, one can find a myriad of various forecasting models that have been developed for the past two decades, and they can be generally classified into three groups:
\begin{itemize}
    \item physical models that are usually based on numerical prediction models
    \item statistical methods, most of which are intelligent algorithms based on data-driven approaches
    \item hybrid physical and statistical models
\end{itemize}

Here we will briefly reflect on different energy sources which are a natural extension and span geophysics research interest.
\subsection{Solar energy}

The total power of solar radiation reaching Earth is approximatelly $1.37\times10^{17} \ W$ \cite{b18} which implies its huge potential for power harvest.
Modern advances of solar energy begins in 1940 with Godfray Cabot and Massachusetts Institute of technology  \cite{b16}, \cite{b17}. Solar radiation defines how much energy strikes to the earth and subsequently implies resources needed for utilization, planning and designing of
solar power plants, in other words photovoltaic (PV) systems. Sun energy that reaches Earth's atmosphere is made of direct and diffuse radiation which are also known as short and long wavelength radiation. The reasons behind using climatological data to estimate solar radiation in the past were due to unreliability of measuring equipment.

Electric energy is produced when photons strike a PV cell and they get absorbed by the semiconductor material.
Energy of the single photon is described by Planck-Einstein relation
\begin{equation}
    E = h f
\end{equation}
where $E [J]$ is the photon energy, $h[Js]$ is the Planck's constant and $f[Hz]$ is the frequency.

Photons can also reflect off the cell or pass through it. So we can see that produced energy is related to PV module efficiency (capability of cell material to efficiently absorb photons being just one of the factors).
Energy produced by a PV module is thus \cite{b19}:
\begin{equation}
    E =  \eta A  G  
\end{equation}
where $E [kW]$ is the power or instantenous energy produced by PV module, $A [m^2]$ is the area covered by PV module, and $G [W/m^2]$ is the total in-plane solar radiation, while $\eta [\%]$ represents  instantaneous PV module efficiency.


\subsection{Wind energy}

Wind energy is becoming more and more popular worldwide, mainly due to an increase of greenhouse gas emission and a solution that this energy resource offers to cut it efficiently. The global wind report, released by the Global Wind Energy Council (GWEC) in 2017, states that the 2016 world wind power market was more than $54.6 GW$, which amounts of the total global installed capacity to nearly $487 GW$. The countries which led at that time were China, US, Germany and India \cite{b8}.

Essentially, wind energy is kinetic energy, and it can be calculated as
\begin{equation}
    E = \frac{1}{2}mv^2 = \frac{1}{2}(\rho Ax)v^2
\end{equation}
where $m [kg]$ is air mass, $v [m/s]$ air stream in the direction at air entrance of rotor blades, $\rho [kg/m^3]$ air density, $A [m^2]$ is cross-sectional area, and $x [m]$ is a thickness of the parcel.
Wind power, which we want to predict, is the derivative of the wind energy and is calculated as follows,
\begin{equation}
    P_w = \frac{dE}{dt} = \frac{1}{2}\rho A v^3
\end{equation}

The unstable and uncontrollable wind speed massively influences the generation of wind power and as such has implications and a heavy impact on wind turbines control, power systems and micro-grid scheduling, power quality and the balance of supply and load demand. This sets the arguments for accurate and trustworthy forecast of wind speed that can not only provide a reliable basis for wind energy generation and thus conversion, but also reduce the costs of power system operation and maintenance \cite{b9}.

\subsection{Wave energy}

Wave energy is an energy source which varies through the day, month, seasons. To be specific, available power from this source varies in time in a way that is uncontrollable. It may be possible to predict the power of waves and time of occurrence at a particular location a few days ahead of time, but it is not possible to control the waves themselves, therefore wave energy forecasting is of outmost importance.
For the previous two decades wave power has seen increase of interest. Although its the most unrepresented one of all the renewables, it has highest production energy potential compared to other renewable sources of energy. Estimates show a potential for wave energy in the order of several thousand $TWh$ \cite{b6}, along the coastlines of the world. Therefore it represents, a vast potential.

Deployment of wave-energy technologies implies that not only
permitting and regulatory matters should be addressed, but also overcoming technological challenges, like e.g. providing accurate prediction of energy generation.



        
To characterize the wave energy, wave power density (WPD) is used, but it should be noted that power output is a nonlinear function of the wave height and period.
The WPD also called wave energy flux (WEF) corresponds to a power content per unit of surface of the crest length.
Wave power density calculation method is as follows for the shallow waters:
\begin{equation}
    P_w = \frac{\rho g}{16} H^2_s\sqrt{gd}
\end{equation}
and for the deep waters we use the following approximation:
\begin{equation}
    P_w = \frac{\rho g^2}{16} H^2_s T_e
\end{equation}
where $P_w[kW/m^2]$ is wave power density,$\rho$ the density
of seawater, $g[m/s^2]$ gravitational acceleration constant,  $H_s[m]$ is the significant wave height, $T_e[s]$ is the energy period and $d[m]$ is the water depth.

\subsection{Oil and gas}

Global fossil fuel energy resources include oil, gas and coal. In 1886, Daimler invented the internal combustion engine, stimulating a great increase in the demand for oil and gas but still coal was dominate energy resource. Advances in geological theory, drilling, and refining technologies helped oil and gas production to increase substantially. Correspondingly, that drove the increase of share of oil and gas in energy sector which grew to more than $50\%$ in 1965. At that point of time coal was replaced as the most utilized energy resource in the world, and with that defining the beginning of the second energy transformation, from coal to oil and gas. The past 10 years, have seen steady growth of oil production, while natural gas production has increased rapidly. 

Crude oil is a mixture of relatively volatile liquid hydrocarbons (compounds which consist mainly of hydrogen and carbon), but it also contains nitrogen, sulfur, and oxygen. Natural gas contains many different compounds and the largest in percentage of components of natural gas is methane. When oil or gas is used as a source of energy, what actually happens is conversion from chemical energy (energy stored in the bonds of atoms and molecules) to thermal energy. The general equation for a complete combustion reaction in which the thermal energy is the main product and radiation (light) side product is
\begin{equation}
    hydrocarbon + O_2 \longrightarrow CO_2 + H_2O
\end{equation}

Oil and gas industry primary concern is  exploration, extraction, refining, transportation (often by oil tankers and pipelines), and marketing of petroleum products (mainly oil and gas). Out of these, we will primarily focus on exploration applications of machine learning.

\section{Paper statistics}
\begin{figure}[htbp]
\centerline{\includegraphics[scale=0.4]{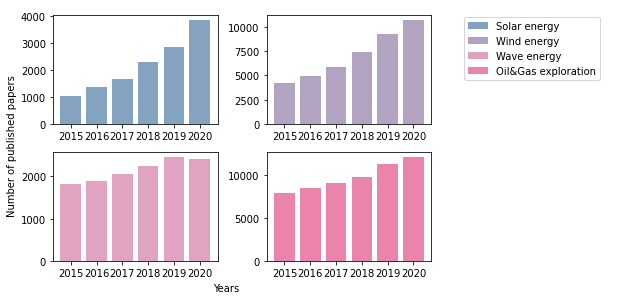}}
\caption{ML algorithms classification}
\label{fig3}
\end{figure}

According to our search results, we have made statistics for number of published papers regarding applications of machine learning algorithms in solar, wind, wave energy forecasting and oil and gas exploration from 2015 to end of 2020.
Figure \ref{fig3} shows statistical results obtained via Google Scholar. The type of papers we searched was limited to research papers and conference
papers.

We can clearly observe the rise of interest in the last five years, especially in the solar energy and wind forecasting, and oil and gas exploration. It is not a surprise that wind energy forecasting and gas and oil exploration lead with the number of published papers, but their rising trend suggest there is still lots of unknowns in these fields. In the next chapter we will try to present the most notable papers with respect to numbers presented here.

\section{ML applications}

\subsection{Solar energy}
Solar energy prediction can be categorized into five types \cite{mla1}:
\begin{itemize}
    \item intra-hour – predicting for next 15 min to 2 hr with a time step of 1 min
    \item hour-ahead – predictions with hourly granularity with a maximum lookahead time of 6 hr
    \item day-ahead – one to three days ahead of hourly predictions;
    \item medium-term – from 1 week to 2 months lead-time and daily production; and
    \item long-term – predicting one to several years for monthly or annual production.
\end{itemize}

An accurate short-term output power forecast of PV systems in power grids or microgrids plays a key role for efficient, economic, stable and sustainable operation of the power supply. So to be more specific, intra-hour prediction is used for forecasting of ramps and high-frequency changes in energy production \cite{mla4}.

\cite{mla2} proposed a solar power prediction model based on a large amount of various historical satellite images and an SVM learning scheme. They've analyzed 4 years of satellite images data of South Korea multiple sites and utilized them with SVM learning. Their proposed SVM-based prediction model can simultaneously predict both the future amount of clouds and solar irradiance in the range of 15 to 300 minutes.

The hybrid model of \cite{mla3} was developed by incorporating the interactions of the PV system supervisory control and data acquisition (SCADA) actual power record with Numerical Weather Prediction (NWP) meteorological data for one year with a time-step of one hour. This approach, a combination of SVM, particle swarm optimization (PSO) and wavelet transform (WT) has a mean computation time smaller than 15 seconds. 
The same year, work published by \cite{mla4} has shown that that ensemble trees (ET) and random forests (RF) performed marginally better than the  support vector regression (SVR). They have also demonstrated that ET has significantly lower training and prediction time, i.e. 8.46 s as compared to 21.5 s and 14 s for RF and SVR, respectively. The paper also proposed using tree-based ensemble methods to provide insight into the analysis of the importance of each input variable as main factors which affect the prediction accuracy.

Paper which employs a hybridized deep learning framework that integrates the convolutional neural network (CNN) for
pattern recognition with the long short-term memory network (LSTM) for half-hourly global solar radiation (GSR) forecasting was proposed by \cite{mla5}. The CNN was applied to robustly extract data input features from predictive variables (i.e., statistically significant antecedent inputs) while LSTM uses them for prediction. 

The work of \cite{mla6} proposes a systematic framework for generating probabilistic forecasts for PV power generation. Their research was based on empirically obtained observation which shows that PV power forecast errors do not follow common distributions like Gaussian, Beta, etc. Therefore, to avoid restrictive assumptions on the shape of the forecast densities, they have proposed a nonparametric density forecasting method based on extreme learning machines (ELM) as a regression tool, trained with an appropriate criterion. The proposed nonparametric method was successfully applied on the two PV power datasets with one-minute resolution and highly fluctuating patterns. Their method efficiently provides reliable and sharp predictive densities for the very short-term (10-minute
and one-hour lead times).

Day ahead and medium-term horizon PV power forecasts, a few hours to days ahead, are used by power grid operators for unit commitment, determining reserve requirements, contingency analysis, and energy storage dispatch.
\cite{mla7} investigated and evaluated applicability of five machine learning models with respect to seasonal effects,namely FoBa,
leapForward, spikeslab, Cubist and bagEarthGCV in modelling solar
irradiance prediction. Main contribution of their work is performance comparison of models in different forecasting horizons ranging from 1 h ahead to 48 h ahead. 
Ensemble approach for day ahead prediction of a PV module power was used by \cite{mla8} where they have used Numerical Weather
Prediction (NWP) input data with different machine learning models and combined them in ensembles to gain better performance and accuracy.

While medium term prediction is used for planning and asset management, long term predictions are useful for assessing resources and selecting potential renewable energy sites \cite{mla10}.
A combination of support vector machines (SVMs) and geographic information systems (GIS) to estimate the rooftop solar PV potential for the urban areas at the commune level was researched in \cite{mla9}. Their approach was to estimate weather variables using SVR which would then constitute inputs of physical models used to estimate and quantify solar energy potential. \cite{mla11} develops an evolutionary seasonal decomposition least-square support
vector regression (ESDLS-SVR) to forecast monthly solar power output. They have also employed genetic algorithms (GA) to select the parameters of the LS-SVR on which  ESDLS-SVR method was built.

\subsection{Wind energy} 

Variability of wind generation can be viewed at various time scales but according to \cite{mla13} falls into these four categories:
\begin{itemize}
    \item very short term ( seconds to 1 hr ahead) 
    \item short term (1–6 hr ahead)
    \item medium term (6–72 hr ahead)
    \item long term (72 hr to years ahead)
\end{itemize}

\cite{mla14} evaluates the performance of eight types of regression trees (RTs) algorithms, including linear and nonlinear approaches of RTs in a real problem of very short-term wind speed prediction from measuring data in wind farms. They have shown that RTs present a small computational burden, which makes them easy for the
retraining of the algorithm in presence of new wind data.
A hybrid model based on autoregressive (AR) model and Gaussian process regression
(GPR) for probabilistic wind speed forecasting was used in \cite{mla15}. In the proposed approach, the AR model is employed to capture the overall structure from wind speed series, and the GPR was adopted to extract the local structure. Additionally, automatic relevance determination (ARD) was used to take into account the relative
importance of different inputs. It should be noted that different types of covariance functions were combined to capture the
characteristics of the data. The proposed hybrid model was compared with the persistence model, artificial
neural network (ANN), and support vector machine (SVM) model for one-step ahead forecasting, using wind
speed data. They have shown that this approach can improve point forecasts compared with other methods, but also generate satisfactory prediction intervals.
Method proposed by \cite{mla16} is based on transferring the information obtained from data-rich farms to a newly-built farms. Their method utilizes deep learning to extract a high-level representation of raw data, to extract wind speed patterns, and then finely tune the mapping with data coming from newly-built farms. Core of this approach is shared-hidden-layer DNN architecture in which the hidden layers are shared across the domains and the output
layers are different in each domain. Their results demonstrate that when there is not lots of training data, DNN models may perform worse than other shallow models, such as SVR and ELM. When data is sufficient, it is only effective when combined with unsupervised pre-training and supervised finetuning strategies.

Study of \cite{mla17} proposes a hybrid forecasting approach that consists of the empirical wavelet transform (EWT), coupled simulated annealing (CSA) and least square support vector machine (LSSVM) for
enhancing the accuracy of short-term wind speed forecasting. The EWT is employed to extract true information from a short-term wind speed series, and the LSSVM, which optimizes the parameters using a CSA algorithm, is used as the predictor to provide the final forecast. Moreover, this study uses a rolling operation method in the prediction processes, including one-step and multi-step predictions, which can adaptively tune the parameters of the LSSVM to respond quickly to wind speed changes. The proposed
hybrid model demonstrated forecasting wind speed series half-hour ahead in the future.
The work of \cite{mla12} first defines which meteorological data need to be included in the predictor, choosing the appropriate weather factors, namely spatially averaged wind speed and wind
direction. Using these inputs, they apply
random forest method to build an hour-ahead wind power predictor. 
The model \cite{mla18} combines
ELM with improved complementary ensemble empirical mode decomposition with adaptive noise (ICEEMDAN) and autoregressive integrated moving average (ARIMA). The ELM model is employed to obtain short-term wind speed predictions, while the autoregressive model is used to determine the best input variables. An ensemble method is used to improve the robustness of the extreme
learning machine. To improve the prediction accuracy, the ICEEMDAN-ARIMA method was developed to postprocess the errors but this method can also be used to preprocess original wind speed. The work of \cite{mla19} proposes hybrid wind speed forecasting model, which includes fast ensemble empirical mode decomposition, sample entropy, phase space reconstruction and back-propagation neural network with two hidden layers, with the aim to enhance the accuracy of wind speed prediction. The data was preprocessed by fast ensemble empirical mode decomposition and sample entropy. Subsequently, the prediction model called improved back-propagation neural network was built to forecast the sub-series, where inputs and outputs were obtained in accordance to phase space reconstruction.

\cite{mla20} paper utilizes SVR for detecting outliers and combines it with seasonal index adjustment (SIA) and Elman recurrent neural network (ERNN) methods to construct the hybrid models named PMERNN and PAERNN. ). First, raw wind speed datasets utilized for model construction were pre-processed by detecting and eliminating
outliers applying the SVR technique. Then, the Kruskale-Wallis (KeW) test was performed to verify the similarity of the pre-processed input datasets distribution. The SIA method was then utilized to extract the seasonal effects from the pre-processed training datasets, while the ERNN model was utilized for training and forecasting trend components. Finally, the predictive trend components of the daily wind speed series were adjusted with the seasonal
index to obtain wind speed forecasts over the prediction horizon.
\cite{mla21} proposed a method based on several ML algorithms to forecast wind
power values efficiently. They have conducted several case studies to investigate performances of Lasso regression, SVR, kNN, xgBoost, RF which showed that machine learning algorithms could be used for forecasting long-term wind power values with respect to historical wind speed data. This study also showed that machine learning-based models could be applied to a location different from model-trained locations demonstrating that machine learning algorithms could be successfully used before the establishment of wind plants in an unknown geographical location.
\cite{mla22} model adopts
heterogeneous base learners and datasets with different resolutions to guarantee diversity. According to the paper,
multi-resolution ensemble scheme, the rapid fluctuation of original high-resolution data could be effectively considered, while a long prediction time scale was also achievable. A total of 16 sub-models were developed to generate preliminary forecasting results. To make the forecasting model adaptive and fit for different datasets, butterfly optimization algorithm (BOA) was employed to select these sub-models, and minimal-redundancy maximal-relevance (mRMR) criterion was set as optimization objective. The selected sub-models were combined by
support vector regression (SVR) to obtain final forecasting result.

\subsection{Wave energy} 

The power produced by ocean wave energy resources is expected to vary unpredictably over timescales ranging from seconds to days and with that in mind we find that the forecast timescale can be categorized as \cite{mla23}:
\begin{itemize}
    \item short term horizons ( 15 min to 6 hours)
    \item medium term horizons ( 6 hours to 24 hours)
    \item long term horizons ( 24 hours to few days)
\end{itemize}

Regarding short term horizons and forecasting of wave height, \cite{mla27} develops a hybrid empirical model decomposition (EMD) support vector regression (SVR) model designated as EMD-SVR for nonlinear and non-stationary wave prediction. Auto-regressive
(AR) model, single SVR model and EMD-AR model were studied to validate the performance of the proposed model. The wavelet decomposition based SVR (WD-SVR) and EMD-SVR models have been investigated to compare the performances of the EMD and WD techniques. In the paper \cite{mla28} , real-time prediction of significant
wave heights for the following 0.5—5.5 h was provided, using information from 3 or more time points. Predictions were made using two ML methods artificial neural networks (ANN) and support vector machines (SVM).

\cite{mla24} paper presents a novel approach for feature selection problems, applied to significant wave height prediction and wave energy flux in oceanic buoys, with marine energy focus. The proposed systems consists of a hybrid grouping genetic algorithm and extreme learning machine (GGA-ELM). It is a wrapper approach, in which the GGA looks for several subsets of features important to solve the problem, and the ELM provides estimation of wave height and wave energy flux prediction in terms of the features selected by the GGA. The novelty of their proposal is that a GGA is used to obtain a reduced number of features for this prediction problem. Similarly, approach \cite{mla25} uses ELM but here the ELM model was coupled with an improved complete ensemble empirical mode decomposition method with adaptive noise (ICEEMDAN) to design
the proposed ICEEMDAN-ELM model. This model incorporates the historical lagged series of Hs as the model's predictor to forecast future significant wave height. The ICEEMDAN algorithm demarcates the original data, into decomposed signals i.e., intrinsic mode functions (IMFs) and a residual component. After decomposition, the partial autocorrelation function is determined for each IMF and the residual sub-series to determine the statistically significant lagged input dataset. The ELM model is applied for forecasting of each IMF by incorporating the significant antecedent significant wave height sub-series as inputs. Finally, all the forecasted IMFs are summed up to obtain the final forecast. The results were benchmarked with those from an online sequential extreme learning machine (OSELM) and random forest (RF) integrated with ICEEMDAN, i.e., the ICEEMDAN-OSELM and ICEEMDAN-RF models. The work of \cite{mla26} uses sequential machine learning approach with the aim to forecast regional waves height. Compared to approaches that involve batch learning algorithms that are not well-equipped to address the demands of continuously changing data stream, their paper conducted a study to predict the daily wave heights in different geographical regions using sequential learning algorithms, namely the Minimal Resource Allocation Network (MRAN) and the Growing
and Pruning Radial Basis Function (GAP-RBF) network.

\cite{mla29} evaluates the accuracy of probabilistic forecasts of wave
energy flux from a variety of methods, including unconditional and conditional kernel density estimation, univariate and bivariate autoregressive moving average generalised autoregressive conditional heteroskedasticity (ARMA-GARCH) models, and a regressionbased method. The bivariate ARMA-GARCH models were implemented with different pairs of variables, such as wave height and wave period, and wave energy flux and wind speed. Their empirical analysis used hourly data from the FINO1 research platform in the North Sea to evaluate density and point forecasts, up to 24 h ahead, for the wave energy flux.

Paper \cite{mla30} uses bayesian optimisation (BO) to obtain the optimal parameters of a prediction system for problems related to ocean wave features prediction. They propose the Bayesian optimization of a hybrid grouping genetic
algorithm (GGA) for attribute selection combined with an extreme learning machine (GGA-ELM) approach for prediction. The system uses data from neighbor stations (usually buoys) in order to predict the significant wave height and the wave energy flux at a goal marine structure facility.
\cite{mla31} approach was to use machine learning models and train them to act as a surrogate for the physics-based SWAN model.
They have used multi-layer perceptron (MLP) model to represent the significant wave height and an SVM model simulated the characteristic period of the wave. The aim of this paper was to show reduction of computational time when machine learning was employed to make forecast compared to physics-based SWAN approach.

\subsection{Oil and gas}

A hydrocarbon is an organic chemical compound composed exclusively of hydrogen and carbon atoms. Hydrocarbons are naturally-occurring compounds and form the basis of crude oil, natural gas, coal. Facies and fractures are the most fundamental features of any geologic formation, which can influence hydrocarbon production. Hence, knowledge of facies and fractures in rocks is critical in oil and gas exploration, and by that we mean identification and prediction of facies and fractures as a means for hydrocarbon exploration. 

One of the paper \cite{mla32}  motivation  was that access to types of high-resolution “geologic ground-truth” data (advanced well log data, image logs, multi-component sonic logs) is limited due to their cost and time related to acquisition so they have tried to compensate that with machine learning approach. They have used bayesian network theory (BNT) to learn the petrophysical data pattern associated with different facies and fractures and random forest (RF) to classify facies and fractures in unconventional shale and conventional sandstone, and carbonate reservoirs. This method showed that both facies and fractures can be predicted with high accuracy using limited common well logs. In addition, the BNT identified the complex causal relationship among the input petrophysical parameters and outputs (facies or fracture). A task of facies classification \cite{mla34} presented as an
application of machine learning, namely the
gradient boosting method, was used for rock facies classification based on certain geological features and constrains. Using XGBoost they have also obtained a feature relative importance of the input data. 

The work of \cite{mla33} explored facies analysis with a goal of reducing time for their analysis using machine-learning techniques. For that purpose, they have used state-of-the-art 3D broadband seismic reflection data and they have compared 20 machine learning models, of which a support vector machine with a cubic kernel function proved to be the best choice for this task.
Purpose of seismic clustering
algorithms is to accelerate this process, allowing one to generate interpreted facies for large 3D volumes, and that is exactly the aim of \cite{mla35}. Determining which attributes best quantify a specific amplitude or morphology component is at the core of this paper. They first apply 3D Kuwahara median filter smoothing the interior attribute response and sharpening the contrast between neighboring facies, thereby preconditioning the attribute volumes for subsequent clustering using generative topographic mapping (GTM).

Work of \cite{mla36} is based on the fact that prestack seismic data carries rich
information which in turn could enable to get higher resolution and more accurate facies maps for further seismic facies recognition task. Because identified object changes from the poststack trace vectors to a prestack trace
matrix, effective feature extraction becomes a challenge which this paper tries to solve. They present a data-driven offset-temporal feature extraction approach using the deep convolutional autoencoder (DCAE). DCAE method learns nonlinear, discriminant, and invariant features from unlabeled data after which seismic facies analysis was accomplished through the use of classification or clustering techniques ( k-means or self-organizing maps (SOM)). Using a physical model and field prestack seismic surveys they have demonstrated that their approach offers the potential to significantly highlight stratigraphic and depositional information.

\cite{mla37} explored the feasibility of constraining the SOM facies analysis using stratigraphy information, in the form of sedimentary cycles. The approach of stratigraphy constrained SOM facies map provides more details and they showed  layers that were more likely overlooked on SOM facies maps without imposed constraints.

To supplement missing logging information without increasing economic cost, \cite{mla38} presented approach to generate synthetic well logs from the existing log data. Incapability of the traditional Fully Connected Neural Network (FCNN) ito preserve spatial dependency was the main motivation for using the Long Short-Term
Memory (LSTM) network, variant of recurrent neural network (RNN).  Using this method, it was shown that synthetic logs can be generated from series of input log data, but with consideration of variation trend and context information with depth.

\section{Conclusion}

From the authors point of view, the ultimate goal of ML-based methods in geophysics would be to leverage and support physical models to obtain synergy between the physical theory from domain scientists and the enhanced, data-driven constraints from ML, probability and statistics theory. Although the application of ML in the cases of solar, wind, wave energy and oil and gas exploration is becoming more and more popular, what was also observed is that ML is often and for most of the time applied without physical modeling. A real transformation, we believe, would start if we could develop more of a hybrid modeling approach that combines data driven ML methods with explicit physical models in order to support model explainability, better performance, etc. 

As we have noted, forecasting of solar, wind and wave energy has seen a lots of interest for the past decade, but the main challenges still remain. It is known that statistical methods like machine learning are good for short term forecasts, but for medium and longer horizons, they need a fusion with physical models. Also, explainability as one of the core issues of black box ML models should also be adressed. In terms of computational power, with respect to big data and computational complexity of hybrid models, accelerators ought to be used e.g. Dataflow paradigm. Transfer learning, related to enriching insufficient site specific data through model generalization over data obtained from different sites, could also be potential topic of further research.

With respect to oil and gas exploration future efforts and research, faster ML integration needs hardware acceleration, transfer learning, automated ML, IoT and edge analytics.

\end{document}